\pdfoutput=1  
\documentclass{article}

\usepackage{arxiv}

\renewcommand{\shorttitle}{Label-Efficient Bilateral Attention for PD Screening}

\usepackage[utf8]{inputenc}
\usepackage[T1]{fontenc}
\usepackage{amsmath,amssymb}
\usepackage{graphicx}
\graphicspath{{./}{figures/}}
\usepackage{booktabs}
\usepackage{array}
\usepackage{xcolor}
\usepackage{url}
\usepackage[hidelinks]{hyperref}

\title{Label-Efficient Bilateral Attention for Parkinson's Disease Screening from Wrist-Worn IMU Signals}

\author{Meheru Zannat \\
	Dept.\ of Computer Science and Engineering \\
	Khulna University of Engineering \& Technology (KUET) \\
	Khulna, Bangladesh \\
	\texttt{zannat2007039@stud.kuet.ac.bd}}

\date{}

\begin{document}
\maketitle

\begin{abstract}
Parkinson's disease (PD) is a chronic neurodegenerative disorder. It shows multiple motor symptoms such as tremor, bradykinesia, postural instability, and freezing of gait (FoG). PD is currently diagnosed clinically through physical examination by health-care professionals, which can be time-consuming and highly subjective. Wearable IMU sensors have become a promising gateway for PD detection. We propose a time-interleaving encoder that processes bilateral wrist-worn IMU signals from the public PADS dataset, which consists of three groups, PD (Parkinson's Disease), HC (Healthy Control), and DD (Differential Diagnosis), across a total of 469 subjects. The encoder reaches $93.2\%$/$90.9\%$ accuracy ($0.963$/$0.960$ AUROC) on HC-vs-PD and PD-vs-DD; the lower sensitivity on the differential ($0.812$ vs.\ $0.994$) confirms that separating PD from look-alike disorders is the harder clinical problem. Self-supervised representation learning with a contrastive InfoNCE objective attains $91.6\%$/$89.5\%$ accuracy using only $20\%$ of the labels, within $1.5$ points of the full-label ceiling, so near-saturation accuracy is reachable with minimal annotation. Beyond the headline numbers we report a systematic ablation of the design space. The bilateral fusion mechanism turns out not to be load-bearing: interleaved self-attention, explicit cross-attention, late concatenation and even single-wrist encoders agree to within $0.001$ AUROC, and a bilateral-pairing control shows the models nonetheless use both streams. What does separate model families is cross-task generalization, where attention encoders hold $0.94$--$0.96$ AUROC on unseen motor tasks while a 1D-CNN collapses to $0.63$. Signal conditioning dominates preprocessing, a direct three-class head is far from ceiling ($89.0\%$ overall, with the other-disorders class hardest at $80.8\%$), and the model runs in $124.5$~ms per window on a Raspberry~Pi~4.
\end{abstract}


\section{Introduction}
Parkinson's disease (PD) is among the most common progressive neurodegenerative disorders, affecting more than 10 million people worldwide~\cite{dorsey2018}. It is incurable, driven by progressive degeneration of the dopaminergic system of the brain~\cite{bloem2021}. It shows several motor symptoms such as tremor, bradykinesia, rigidity, and postural instability. These continuous, episodic symptoms can be captured passively by wrist-worn smartwatches with embedded inertial measurement unit (IMU) sensors.
A particular clinical difficulty is that these signs are shared with other conditions grouped as differential diagnoses (DD), such as essential tremor, atypical Parkinsonism, and multiple sclerosis, and can lead
to misdiagnosis and delayed treatment at an early stage. Yet most wearable studies stop at PD-versus-healthy detection; the clinically decisive PD-versus-DD boundary is rarely addressed.

We target both decisions on the public PADS dataset~\cite{varghese2024}, which
provides synchronized left- and right-wrist IMU recordings. Motor asymmetry, where
signs onset and progress more severely on one side, is a hallmark of PD, and it
motivates encoding both wrists jointly rather than in isolation. Instead of
hard-wiring the bilateral interaction, we embed the two wrists into a single sequence
and let self-attention discover it: attention over the concatenated stream computes
within-wrist and between-wrist interactions in one operation with a single tied set of
weights. We evaluate this design against an explicit dual-stream cross-attention
encoder, late feature concatenation, single-wrist ablations, and 1D-CNN and BiLSTM
baselines under a patient-disjoint protocol, reporting threshold-free AUROC
throughout. Our contributions are:
\begin{itemize}
\item A compact interleaved self-attention encoder with a hierarchical two-head
formulation (HC-vs-PD screening and PD-vs-DD differential diagnosis) that outperforms a
direct three-class classifier and targets the clinically challenging PD-vs-DD boundary.
\item A systematic ablation of the design space (bilateral fusion mechanism, encoder
family, and band-pass filtering) that identifies which components actually drive
in-distribution accuracy and cross-task generalization.
\item A contrastive self-supervised pre-training scheme that outperforms supervised
training at low label budgets and matches it beyond ${\sim}20\%$ of labels, cutting the
annotation cost.
\item An inference rate of $124.5$~ms per window on a Raspberry~Pi~4 CPU with minimal
preprocessing, establishing the feasibility of on-device edge deployment.
\end{itemize}

\section{Related Work}
\label{sec:related}
Objective of PD detection with wearable IMU sensor modality is being extensively investigated recently.

\begin{table}[!b]
\caption{Comparison of related work. $\checkmark$=addressed;
         $\times$=not addressed; $\sim$=partial.}
\label{tab:related_work}
\begin{center}
\begin{tabular}{llccc}
\toprule
\textbf{Work} & \textbf{Method}
  & \textbf{\begin{tabular}[c]{@{}c@{}}Label\\Efficiency\end{tabular}}
  & \textbf{\begin{tabular}[c]{@{}c@{}}PD vs.\\DD\end{tabular}}
  & \textbf{\begin{tabular}[c]{@{}c@{}}Edge\\Deploy\end{tabular}} \\
\midrule
Patel et al.~\cite{patel2009}          & Handcrafted features          & $\times$ & $\times$ & $\times$ \\
Sigcha et al.~\cite{sigcha2022}        & CNN-Transformer               & $\times$ & $\times$ & $\times$ \\
Guo et al.~\cite{guo2022}              & LSTM                          & $\times$ & $\times$ & $\times$ \\
Zhao et al.~\cite{zhao2025mfam}        & Multi-scale freq.\ network    & $\times$ & $\times$ & $\times$ \\
Soumma et al.~\cite{soumma2025liftpd}  & SSL (LIFT-PD)                 & $\sim$   & $\times$ & $\times$ \\
Farahmand et al.~\cite{farahmand2024gluconet} & Hybrid attention (GlucoNet) & $\times$ & $\times$ & $\times$ \\
\midrule
\textbf{Ours} & \textbf{Interleaved self-attention + SSL}
  & $\checkmark$ & $\checkmark$ & $\checkmark$ \\
\bottomrule
\end{tabular}
\end{center}
\end{table}

Most of the earlier works relied on handcrafted features derived from accelerometer signals, to determine tremor and bradykinesia~\cite{patel2009}; CNNs were used for IMU time series analysis of gait and tremor~\cite{soumma2025fog}. While LSTMs have been applied to capture sequential motor signals dependencies~\cite{guo2022}, and CNN-Transformer hybrids for single waist-worn accelerometer data for FoG detection have been employed~\cite{sigcha2022}. A limitation common across this work is the heavy reliance on unilateral sensing, despite the well-known motor asymmetry in PD symptoms.
Transformer architecture~\cite{vaswani2017} is suitable for modelling the irregular, periodic nature of motor symptoms, as self-attention lets the model make associations between distant timestamps simultaneously.

For Human Activity Recognition (HAR) tasks, dual-stream architectures have been designed using parallel encoders for preprocessing various sensor modalities. The PD has also been explored in multi-scale frequency-aware architectures for severity assessment~\cite{zhao2025mfam}. Yet existing dual-stream designs fuse the streams using concatenation or addition, without any explicit modelling of the contralateral relation between left and right motor signals.

Self-supervised contrastive representation learning~\cite{oord2018} has become a powerful learning paradigm for unlabelled clinical data. TS-TCC~\cite{eldele2021} combines temporal and contextual dissimilarity. LIFT-PD~\cite{soumma2025liftpd} gaining competitive performance with $40\%$ fewer labelled samples. These approaches successfully transfer to downstream clinical task but have not been validated on bilateral IMU data for PD classification, where augmentation strategies need to ensure clinically meaningful asymmetric bands between the two wrist streams.

Finally, an continuing challenge in clinical ML is to avoid optimistic bias by conflating model selection with performance estimation~\cite{varma2006}. The vast majority of studies reporting PD detection evaluate results from a single split or flat $k$-fold; they yield models that are difficult to compare, and potentially too optimistic.

All these challenges have been addressed in our studies, which addressed the challenges in detecting the asymmetry in PD symptoms, scarcity of labelled data, generalization and real-life feasibility. Table~\ref{tab:related_work} positions this work against the closest prior studies.

\section{Dataset}

\begin{figure}[!t]
	\centering
	\includegraphics[width=\linewidth]{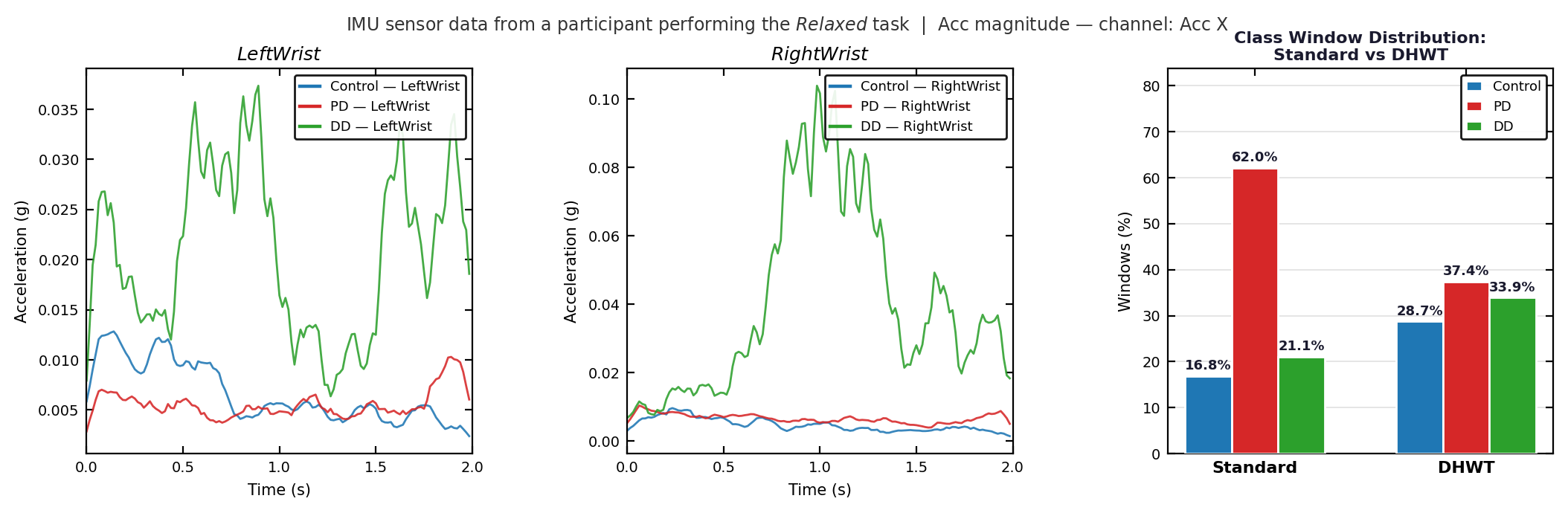}
	\caption{Representative Acc-magnitude signals for the \textit{Relaxed} task from
	Control, PD, and DD participants at the (a)~left and (b)~right wrist. (c)~Class
	balance of training windows under standard non-overlapping windowing versus the
	class-dependent DHWT scheme, which equalizes the three groups.}
	\label{fig:eda}
\end{figure}

PADS~\cite{varghese2024} contains bilateral wrist-worn smartwatch recordings from 469 participants performing ten standardized motor tasks designed by neurologists: 291 with PD, 79 healthy controls (HC), and 99 with differential diagnoses (DD). Each participant wears two Apple Watch Series~4 devices, one per wrist, giving six-axis IMU measurements per wrist (three accelerometer axes $a_x,a_y,a_z$ and three gyroscope axes $\omega_x,\omega_y,\omega_z$) sampled at $100$~Hz. The tasks (\textit{CrossArms}, \textit{DrinkGlas}, \textit{Entrainment},
\textit{HoldWeight}, \textit{LiftHold}, \textit{PointFinger}, \textit{Relaxed}, \textit{StretchHold}, \textit{TouchIndex}, \textit{TouchNose}) are each $10$--$20$~seconds long. PADS lacks uniform severity and time-since-diagnosis labels, so we make no claim about severity staging, only categorical detection and differentiation.
Figure~\ref{fig:eda} shows representative signals for each group and
the effect of our class-balancing on the training-window distribution.

\section{Proposed Method}
\label{sec:method}
\begin{figure}[t]
\centering
\IfFileExists{figures/interleaved_attn.pdf}{%
  \includegraphics[width=1.00\linewidth]{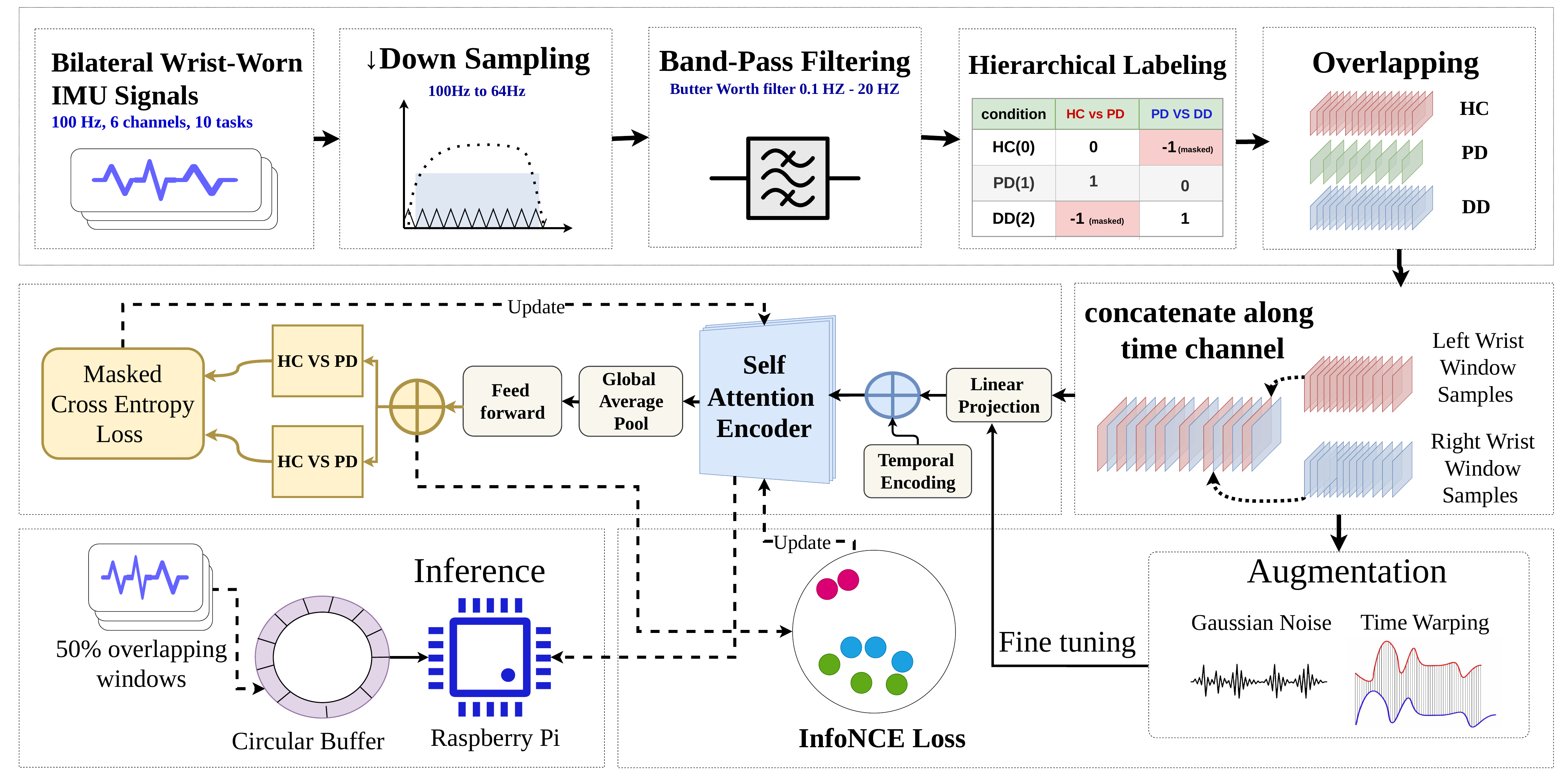}%
}{%
  \fbox{\parbox[c][45mm][c]{0.98\linewidth}{\centering\itshape
  [ figure placeholder --- export \texttt{figures/parkinsons\_interleave.drawio}
  to \texttt{figures/interleaved\_attn.pdf} ]}}%
}
\caption{Interleaved self-attention encoder: the two wrist streams are projected,
tagged with a learned modality embedding, and concatenated along time into a single
$2T$ sequence, over which standard self-attention computes within-wrist (diagonal) and
between-wrist (off-diagonal) interactions in one operation. The same encoder is used
for supervised training, contrastive self-supervised pre-training, and inference.}
\label{fig:arch}
\end{figure}

This section describes only the method we propose. Every design alternative we
considered, and every ablation that justifies a choice made here, is deferred to
Section~\ref{sec:results}. Figure~\ref{fig:arch} shows the complete pipeline: signal
conditioning, hierarchical labelling, class-dependent windowing, the interleaved
encoder, the contrastive pre-training branch, and the on-device inference loop.

\subsection{Problem Formulation}
Each sample is a pair of synchronous left ($L$) and right ($R$) wrist signals
$\mathbf{X}=(\mathbf{X}^L,\mathbf{X}^R)\in\mathbb{R}^{T\times 2C}$, with $C{=}6$ channels per wrist. We decompose the three-way problem into two binary heads: HC-vs-PD (screening) and PD-vs-DD (differential diagnosis). A hierarchical masking scheme supplies each head only the relevant samples: for an HC subject the PD-vs-DD label is masked ($-1$); for a DD subject the HC-vs-PD label is masked; PD subjects are positive for the first head and negative for the second. Masked labels are excluded from the corresponding head's loss. Section~\ref{sec:threeclass} shows that this decomposition outperforms a single three-class head.

\subsection{Data Preprocessing}
The six IMU channels per wrist are retained and the first $0.5$~s ($50$ samples) of
each task removed to discard initialization transients. Signals are down-sampled from $100$~Hz to $64$~Hz by polyphase resampling, then filtered with a fifth-order Butterworth band-pass ($[0.1,20]$~Hz) using zero-phase forward--backward filtering (\texttt{filtfilt}). Clinically relevant activity (resting tremor $4$--$6$~Hz and voluntary movement $\leq 5$~Hz) lies inside this passband, while DC offsets and high-frequency sensor artifacts are removed; the effect is quantified in Section~\ref{sec:bp}.
Signals are segmented into windows of $T{=}256$ timestamps
($4$~s at $64$~Hz). Because the group counts are imbalanced (HC:79, PD:291, DD:99), uniform windowing collapses to the majority class; we therefore apply a class-dependent overlap (Differential Hopping Windowing~\cite{soumma2025liftpd}, $70\%$ for HC, $0\%$ for PD, $65\%$ for DD), yielding ${\sim}8{,}500$ windows per class. We use patient-disjoint stratified $5$-fold cross-validation, so no subject appears in both training and validation.

\subsection{Bilateral Fusion Strategy}
PD signs typically begin on one side and remain more severe there, so the relationship
between the two wrists carries diagnostic information that a single-wrist model
discards. Rather than committing to a hand-designed interaction, we fuse the wrists by
\emph{time interleaving}: both streams are projected into a shared embedding space,
tagged with a learned modality embedding that identifies the wrist, and concatenated
along the time axis into one $2T$ sequence. Ordinary self-attention over that sequence
then computes within-wrist and between-wrist relationships in a single operation under
one tied set of weights, rather than through separate query/key pathways per direction.
Section~\ref{sec:fusion} compares this against explicit dual-stream cross-attention,
late concatenation, and single-wrist inputs.

\subsection{Interleaved Self-Attention Encoder}
\label{sec:encoder}
Each $C{=}6$-channel input is linearly projected to $d$ dimensions with added sinusoidal positional encodings~\cite{vaswani2017} and a learned modality embedding $\mathbf{m}^L,\mathbf{m}^R\in\mathbb{R}^{d}$ identifying the wrist,
then concatenated along the time axis ($\Vert$):
\begin{equation}
\mathbf{H}^{(0)}=\big[\,\tilde{\mathbf{H}}^{L}\;\Vert\;\tilde{\mathbf{H}}^{R}\,\big]
\in\mathbb{R}^{2T\times d},
\end{equation}
where $\tilde{\mathbf{H}}^{w}=\mathrm{PE}(\mathbf{W}^{w}\mathbf{X}^{w})+\mathbf{m}^{w}$
for $w\in\{L,R\}$. The full $2T$ sequence then passes through $N$ standard transformer
blocks. Each block applies multi-head self-attention (MHA) with a residual connection
and layer normalization, followed by a position-wise feed-forward network:
\begin{equation}
\mathbf{H}^{(n)}=\mathrm{FFN}\!\left(\mathrm{LayerNorm}\!\left(
  \mathbf{H}^{(n-1)}+\mathrm{MHA}(\mathbf{H}^{(n-1)})\right)\right).
\end{equation}
Because attention is taken over the concatenated stream, each $2T{\times}2T$ attention map decomposes into four $T{\times}T$ quadrants: the two diagonal blocks are within-wrist self-attention and the two off-diagonal blocks are left--right
cross-attention. The interleaved encoder therefore realizes the same set of bilateral interactions as an explicit dual-stream cross-attention design, but with one tied set of projection weights instead of four separate ones ($159{,}556$ against $418{,}180$ parameters at $d{=}64$) and one attention operation per block instead of four. The two share a single softmax normalization across the $2T$ positions, so left and right
timesteps compete for attention mass, the one substantive functional difference.

Global average pooling over the $2T$ output positions yields $\mathbf{z}\in\mathbb{R}^d$,
passed to two linear~+~softmax heads, one per binary decision.

\subsection{Self-Supervised Pre-training}
To reduce the annotation burden, we pre-train the
encoder with a contrastive objective~\cite{chen2020} and then fine-tune. For each anchor $z_i$, a
positive $z_i^+$ is formed by augmentation and the other in-batch samples are
negatives; \emph{no class labels are used at any point during pre-training}. Two
augmentations alternate with equal probability: cubic-spline temporal
warping~\cite{um2017} and additive Gaussian noise ($\sigma{=}0.05$). The InfoNCE
loss~\cite{oord2018} per batch of size $B$ is
\begin{equation}
\mathcal{L}_{\mathrm{NCE}}=-\frac{1}{B}\sum_{i=1}^{B}
\log\frac{\exp(\mathrm{sim}(z_i,z_i^+)/\tau)}
{\sum_{j=1}^{B}\exp(\mathrm{sim}(z_i,z_j)/\tau)}.
\end{equation}
The pre-trained encoder is then either fine-tuned end-to-end or frozen with a linear
probe on top.

\subsection{Optimization}
Cross-entropy is computed per head over
non-masked samples and averaged. We train with AdamW~\cite{loshchilov2019} (weight
decay $0.01$), unit-norm gradient clipping, ReduceLROnPlateau, up to $100$ epochs,
batch size $32$, learning rate $5{\times}10^{-4}$, and dropout $0.2$; the base model
uses $d{=}64$, $N{=}3$ blocks, $h{=}8$ heads, and a $256$-dimensional feed-forward
layer. These settings follow an Optuna (TPE) search over model width, depth, learning
rate, weight decay and dropout, which consistently favoured \emph{small} architectures
($d{=}32$--$64$, $N{=}2$--$3$ blocks), indicating that a compact, well-regularized
model generalizes best on this dataset.

\subsection{Inference and Edge Deployment}
For deployment the offline zero-phase filter is replaced by a causal
second-order-section band-pass, since \texttt{filtfilt} is not realizable in a
streaming setting. Samples arrive at $100$~Hz, are down-sampled to $64$~Hz and filtered
sample-by-sample with persistent filter state. A $256$-sample circular buffer holds the
current window and inference is triggered every $128$ samples ($2$~s step, $50\%$
overlap). The deployed configuration is the $d{=}64$, $N{=}3$ interleaved encoder
running on a Raspberry~Pi~4 (ARM Cortex-A72, 4~GB RAM); measured latency is reported in
Section~\ref{sec:edge}.

\section{Experiments and Results}
\label{sec:results}
All models were implemented in PyTorch on an NVIDIA GPU with SciPy for signal
processing; hyper-parameters were searched with Optuna (TPE). We report window-level
metrics as mean over the patient-disjoint $5$-fold cross-validation. Because both
tasks are close to balanced (HC-vs-PD prevalence $0.52$; PD-vs-DD $0.47$), we
emphasize threshold-independent AUROC alongside accuracy and F1. We note that
patient-level AUROC saturates at $1.000$ for every encoder on both tasks, so all
comparisons below are made at the window level, which is the only level that
discriminates between models.

\subsection{Detection and Differential Diagnosis}
\begin{figure}[t]
\centering
\includegraphics[width=0.85\linewidth,clip,trim=0 0 322 0]{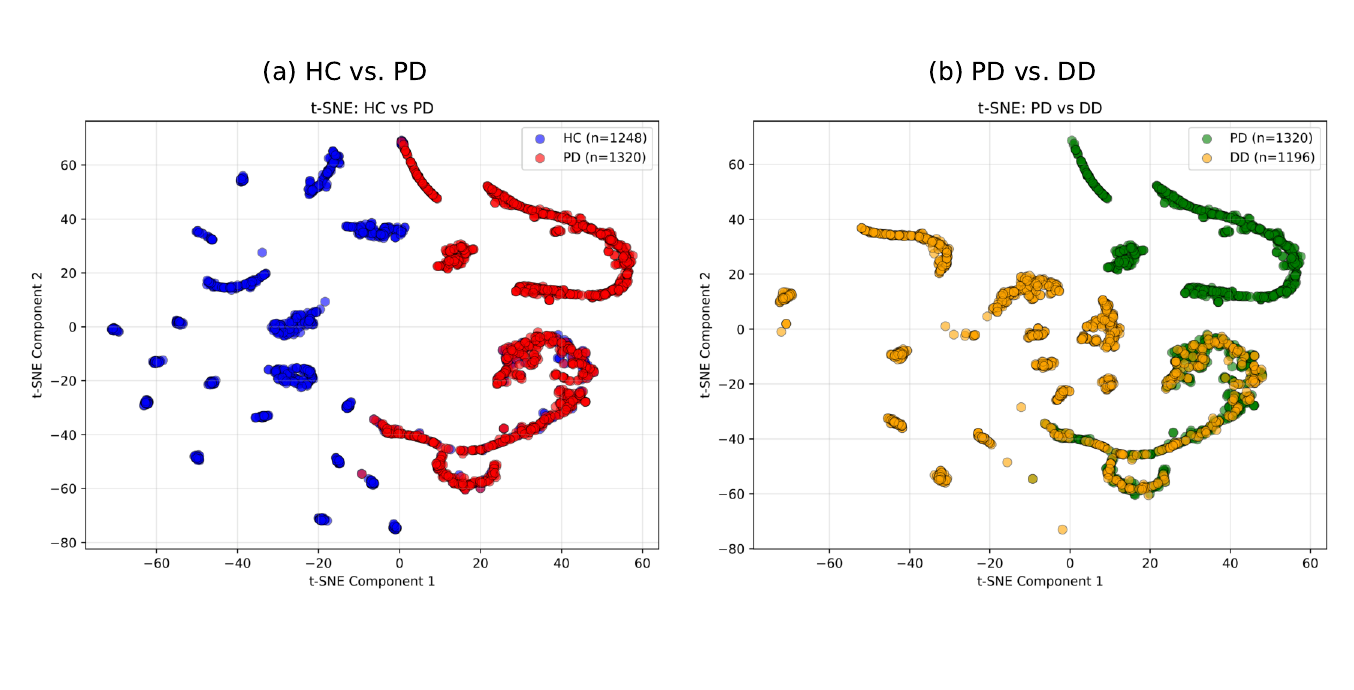}
\caption{t-SNE of the learned embeddings. (a)~HC and PD separate cleanly, whereas
(b)~PD and DD share a large overlapping region---the geometric signature of the harder
differential-diagnosis boundary.}
\label{fig:tsne}
\end{figure}

Table~\ref{tab:main} reports the interleaved self-attention encoder on both decisions.
HC-vs-PD reaches $0.963$ AUROC with near-perfect sensitivity ($0.994$), whereas
PD-vs-DD, at comparable accuracy and equal stability (between-fold std $\leq 0.002$
AUROC), attains lower F1 ($0.894$) and much lower sensitivity ($0.812$). This asymmetry is the central
clinical finding: distinguishing PD from look-alike disorders that share tremor and
bradykinesia is genuinely harder than distinguishing PD from healthy controls, and a
system tuned for screening will under-detect atypical cases unless the differential is
modelled explicitly. The learned embeddings make this geometry explicit
(Fig.~\ref{fig:tsne}): HC and PD separate cleanly, whereas PD and DD share a large
overlapping region.

\begin{table}[t]
\caption{Interleaved self-attention encoder (window-level, with band-pass, $5$
patient-disjoint folds). PD-vs-DD is as stable as HC-vs-PD but shows lower
AUROC/F1/sensitivity---the harder clinical boundary.}
\label{tab:main}
\centering
\begin{tabular}{lccccc}
\toprule
\textbf{Task} & Acc & AUROC & F1 & Sens & Spec \\
\midrule
HC vs.\ PD & 0.932 & 0.963 & 0.938 & 0.994 & 0.865 \\
PD vs.\ DD & 0.909 & 0.960 & 0.894 & 0.812 & 0.996 \\
\bottomrule
\end{tabular}
\end{table}

\subsection{Bilateral Fusion: Does the Mechanism Matter?}
\label{sec:fusion}
Because the encoder is motivated by bilateral sensing, we first ask whether the
\emph{fusion mechanism} is load-bearing. Table~\ref{tab:fusion} compares three ways of
combining the wrists under an identical pipeline---interleaved self-attention over the
$2T$ stream, explicit dual-stream cross-attention, and late concatenation of the two
pooled representations---together with single-wrist ablations.

\emph{No mechanism separates from any other.} Window-level AUROC spans $0.963$--$0.965$
on HC-vs-PD and $0.960$--$0.961$ on PD-vs-DD; the largest gap between any two fusion
variants is $0.001$ AUROC, an order of magnitude smaller than the between-fold
variation, and paired bootstrap tests over patients place every pairwise difference
inside its $95\%$ confidence interval. Single-wrist encoders match the bilateral ones
to the same tolerance. We therefore make no accuracy claim for bilateral fusion, and
adopt the interleaved variant on efficiency grounds alone.

That the mechanism does not change accuracy does not mean the models ignore the second
wrist. A \emph{bilateral-pairing control}---re-pairing each left-wrist window with a
right-wrist window drawn from a different sample, leaving both signals otherwise
intact---collapses the interleaved encoder from $0.962$/$0.960$ AUROC to
$0.622$/$0.586$, and cross-attention from $0.963$/$0.960$ to $0.561$/$0.556$. Late
concatenation, which mixes the streams only after pooling, is the least disrupted
($-0.15$ AUROC), and a left-wrist-only encoder is by construction unaffected
($\Delta=0.000$), serving as a positive control on the procedure. The bilateral models
thus genuinely integrate the two streams and break when the correspondence is
destroyed---the information is redundant for this classification, but it is being used.

\begin{table}[t]
\caption{Bilateral fusion mechanisms and single-wrist ablations (window-level,
$5$ patient-disjoint folds; parameters at $d{=}64$). All mechanisms agree to within
$0.001$ AUROC; we adopt the interleaved encoder for its lower parameter and compute
cost, not its accuracy.}
\label{tab:fusion}
\centering
\begin{tabular}{lccccc}
\toprule
& & \multicolumn{2}{c}{\textbf{HC vs.\ PD}} & \multicolumn{2}{c}{\textbf{PD vs.\ DD}}\\
\cmidrule(lr){3-4}\cmidrule(lr){5-6}
\textbf{Encoder}& Params & Acc & AUROC & Acc & AUROC \\
\midrule
Interleaved self-attn (ours) & \textbf{159.6\,K} & 0.932 & 0.963 & 0.909 & 0.960 \\
Cross-attention              & 418.2\,K & 0.926 & 0.964 & 0.909 & 0.960 \\
Late concatenation           & 317.6\,K & 0.928 & 0.964 & 0.909 & 0.960 \\
Left wrist only              & 159.4\,K & 0.928 & 0.964 & 0.909 & 0.960 \\
Right wrist only             & 159.4\,K & 0.929 & 0.965 & 0.909 & 0.961 \\
\bottomrule
\end{tabular}
\end{table}

\subsection{Encoder Comparison and Cross-Task Generalization}
\label{sec:crosstask}
\begin{figure}[t]
\centering
\includegraphics[width=0.85\linewidth]{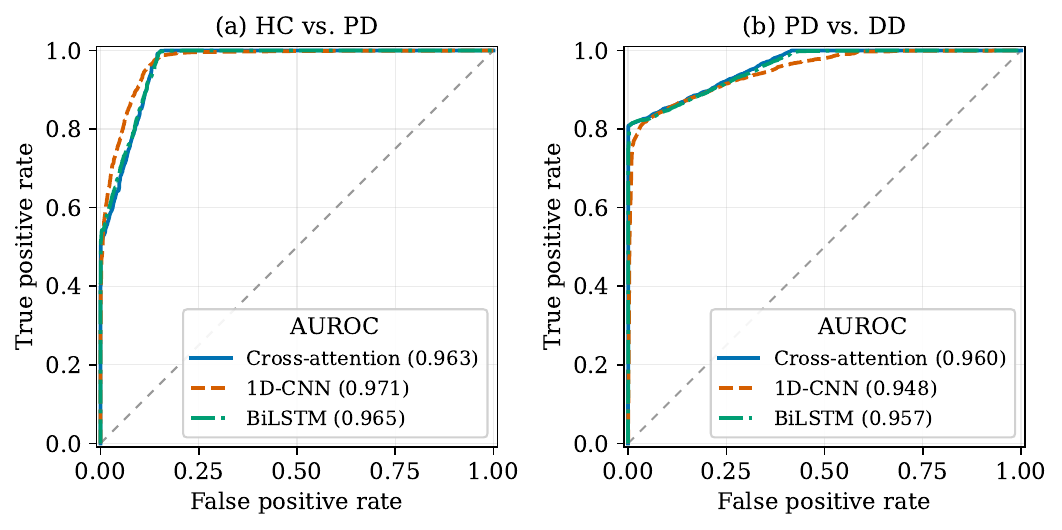}
\caption{Comparative ROC curves (window-level, pooled over the five patient-disjoint
folds). In distribution the encoders are nearly identical; the 1D-CNN even edges the
attention encoder on HC-vs-PD (a) yet is weakest on PD-vs-DD (b).}
\label{fig:roc}
\end{figure}

We next compare the interleaved encoder against 1D-CNN and BiLSTM baselines under
an identical pipeline. \emph{In distribution the three are near-indistinguishable}:
Fig.~\ref{fig:roc} shows heavily overlapping ROC curves, and the 1D-CNN is in fact
marginally the strongest on HC-vs-PD ($0.971$ AUROC) while being the weakest on
PD-vs-DD ($0.948$). In-distribution accuracy therefore cannot tell these models apart
or justify a choice between them.

The difference appears only when we ask a harder, clinically realistic question: does
a model trained on one set of motor tasks generalize to \emph{tasks it has never
seen}? Under leave-one-task-out (LOTO)---holding out each of the ten tasks in turn,
training on the other nine, and averaging AUROC over the ten held-out tasks
(Table~\ref{tab:loto})---the interleaved encoder stays at $0.959$/$0.943$ AUROC
(HC/PD, PD/DD) with a worst held-out task no lower than $0.930$. The 1D-CNN
\emph{collapses} to $0.802$/$0.789$ on average and to $0.631$/$0.540$ on its worst
task (near chance), and the BiLSTM---competitive on average ($0.954$)---is three-to-five
times more variable and drops to $0.900$. The same ranking holds when we instead train
on seven tasks and test on three unseen ones (attention $0.964$/$0.969$; CNN
$0.769$/$0.745$; BiLSTM $0.918$/$0.893$). Attention encoders thus learn task-invariant
motor structure and generalize robustly; the 1D-CNN instead latches onto task-specific
patterns that do not transfer, while the BiLSTM generalizes on average but far less
reliably across tasks. Notably, this separation is driven entirely by the encoder
\emph{family}. Table~\ref{tab:loto} lists all five attention variants: interleaved,
cross-attention, late concatenation, and both single-wrist ablations transfer within
$0.003$ AUROC of one another ($0.958$--$0.960$ on HC-vs-PD, $0.943$--$0.946$ on
PD-vs-DD), while the CNN loses more than $0.15$. The fusion mechanism is therefore
irrelevant to transfer just as it is to in-distribution accuracy; what matters is
attention and compactness, not how---or even whether---the two wrists are combined.

\begin{table}[t]
\caption{Leave-one-task-out generalization (train on nine tasks, test on the held-out
one; AUROC averaged over the ten held-out tasks, worst single task in parentheses).
\emph{Every} attention encoder transfers within $0.003$ AUROC of every other,
including the single-wrist ablations; only the encoder family separates, with the
1D-CNN collapsing and the BiLSTM far more variable.}
\label{tab:loto}
\centering
\begin{tabular}{lcc}
\toprule
\textbf{Model} & \textbf{HC vs.\ PD} & \textbf{PD vs.\ DD} \\
               & mean (worst) & mean (worst) \\
\midrule
\multicolumn{3}{l}{\textit{Attention encoders}}\\
Interleaved self-attn (ours) & 0.959 (0.952) & 0.943 (0.930) \\
Cross-attention              & 0.958 (0.944) & 0.944 (0.929) \\
Late concatenation           & 0.960 (0.950) & 0.943 (0.930) \\
Left wrist only              & 0.960 (0.954) & 0.946 (0.930) \\
Right wrist only             & 0.960 (0.951) & 0.945 (0.929) \\
\midrule
\multicolumn{3}{l}{\textit{Non-attention baselines}}\\
1D-CNN                       & 0.802 (0.631) & 0.789 (0.540) \\
BiLSTM                       & 0.954 (0.900) & 0.933 (0.896) \\
\bottomrule
\end{tabular}
\end{table}

\subsection{Effect of Band-Pass Filtering}
\label{sec:bp}
\begin{figure}[t]
\centering
\includegraphics[width=0.85\linewidth]{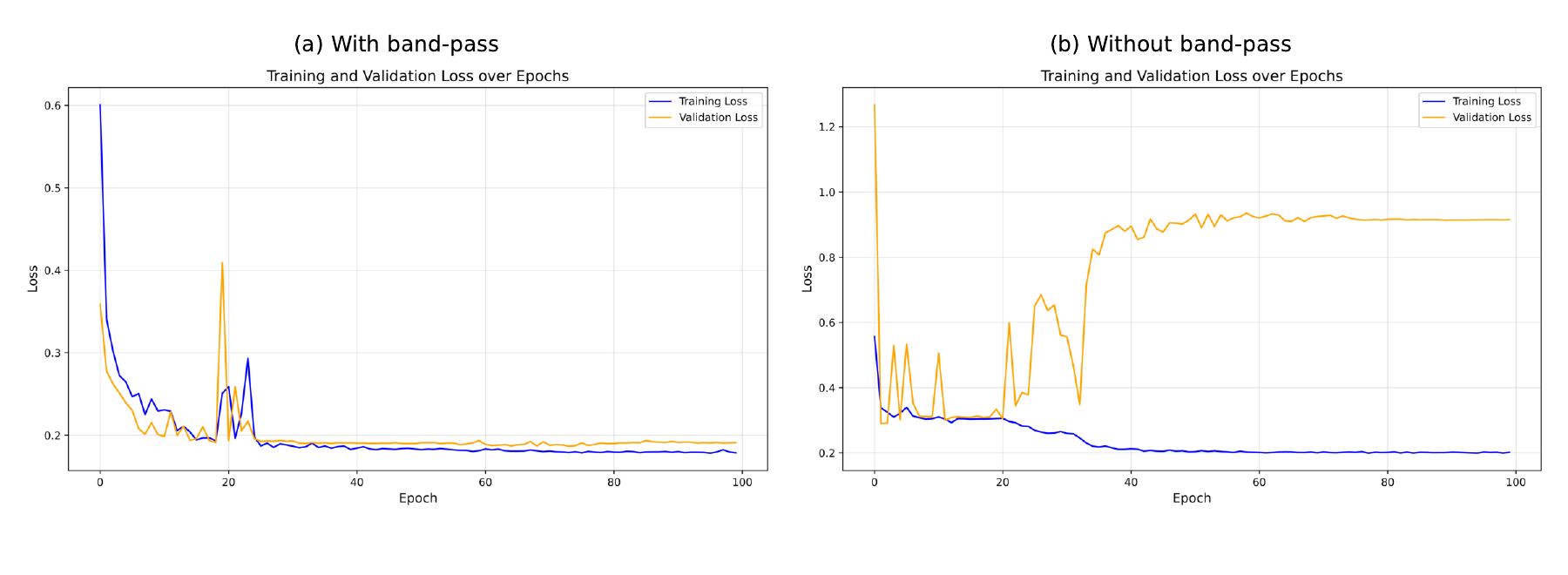}
\caption{Training/validation loss (a)~with and (b)~without the band-pass filter.
Without the filter, validation loss diverges from training loss, the signature of
overfitting to out-of-band noise.}
\label{fig:loss}
\end{figure}

Removing the band-pass filter is the largest single effect we observe. As a
preprocessing ablation it is measured with the encoder held fixed; we report it on the
cross-attention variant, which Table~\ref{tab:fusion} shows is interchangeable with the
interleaved encoder for this purpose. Without the filter, validation loss diverges from
training loss (Fig.~\ref{fig:loss}), indicating that out-of-band noise leads the model
to memorize high-frequency artifacts. Accuracy drops to $87.6\%$ (HC-vs-PD) and
$84.2\%$ (PD-vs-DD) and AUROC falls from $0.964$ to $0.936$ and from $0.960$ to
$0.920$; adding the filter improves accuracy by $+5.0$ and $+6.7$ points and reduces
between-fold standard deviation by roughly an order of magnitude
($0.0148\rightarrow0.0010$ on HC-vs-PD). In distribution this is a far larger and more
consistent effect than either the encoder family or the fusion mechanism, confirming
that the motor-relevant signal lies in the $0.1$--$20$~Hz band.

\subsection{Label Efficiency}
\begin{figure}[t]
\centering
\includegraphics[width=0.85\linewidth]{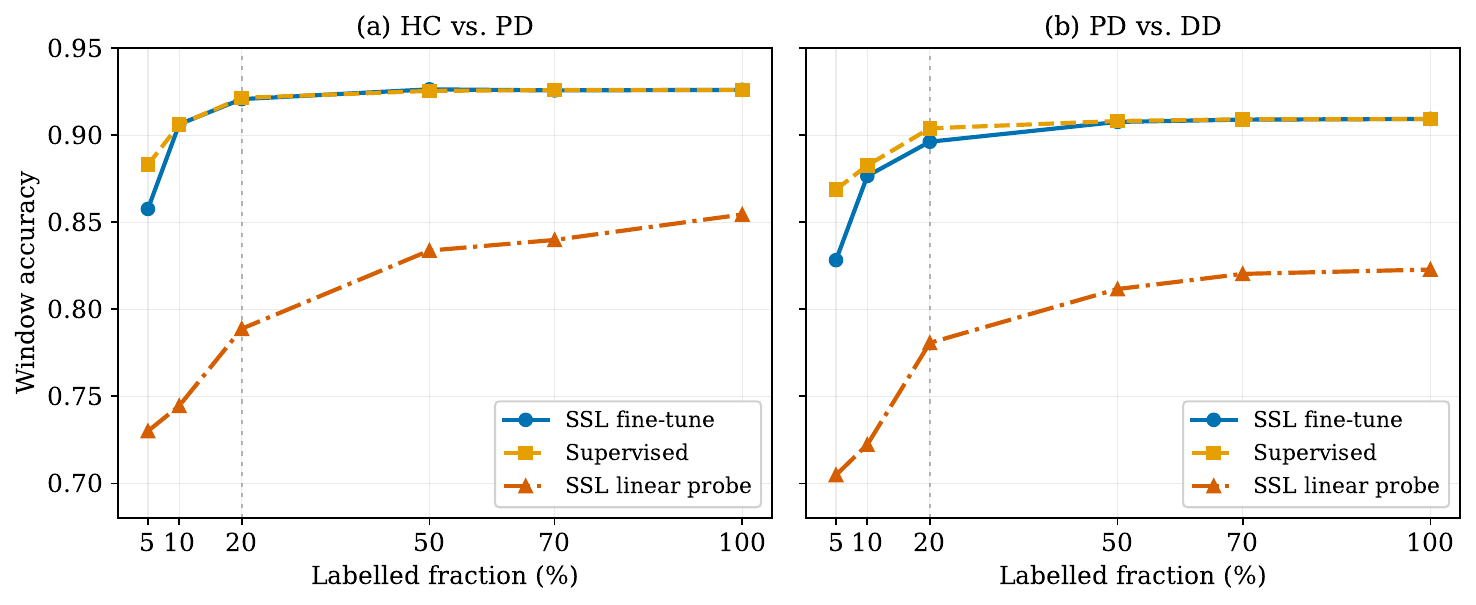}
\caption{SSL label-efficiency curves: window accuracy versus labelled fraction for
(a)~HC-vs-PD and (b)~PD-vs-DD. Contrastive pre-training leads clearly in the low-label
regime and both curves plateau near ${\sim}20\%$ of labels (dashed line), after which
they are indistinguishable.}
\label{fig:labeleff}
\end{figure}

Table~\ref{tab:labeleff} and Figure~\ref{fig:labeleff} compare SSL fine-tuning against
supervised training from scratch at increasing labelled fractions. The task is highly
label-efficient: both regimes reach within ${\sim}1$ point of their full-label accuracy
using only ${\sim}20\%$ of labels. \emph{Pre-training pays off precisely where labels
are scarce}: at a $5\%$ budget SSL fine-tuning leads supervised training by $9.8$
points on HC-vs-PD ($0.849$ vs.\ $0.751$) and $10.5$ points on PD-vs-DD ($0.832$ vs.\
$0.728$), and still leads at $10\%$; from $20\%$ onward the two are indistinguishable.
Much of this gap is stability rather than ceiling---at $5\%$ the supervised runs vary
enormously across folds (std $0.108$) whereas the pre-trained ones do not (std
$0.025$), so contrastive pre-training mainly supplies an initialization from which
low-label training reliably converges. This is the regime that matters clinically,
where expert-labelled recordings are the binding constraint.

\begin{table}[t]
\caption{Label efficiency (window-level accuracy, $5$-fold mean, interleaved encoder).
SSL fine-tuning leads clearly at $5$--$10\%$ of labels and matches supervised training
from $20\%$ onward; both plateau near ${\sim}20\%$.}
\label{tab:labeleff}
\centering
\begin{tabular}{lcccc}
\toprule
& \multicolumn{2}{c}{\textbf{HC vs.\ PD}} & \multicolumn{2}{c}{\textbf{PD vs.\ DD}}\\
\cmidrule(lr){2-3}\cmidrule(lr){4-5}
\textbf{Labels} & SSL FT & Superv. & SSL FT & Superv. \\
\midrule
$5\%$   & \textbf{0.849} & 0.751 & \textbf{0.832} & 0.728 \\
$10\%$  & \textbf{0.896} & 0.878 & \textbf{0.873} & 0.832 \\
$20\%$  & 0.916 & 0.921 & 0.895 & 0.901 \\
$50\%$  & 0.925 & 0.926 & 0.907 & 0.909 \\
$100\%$ & 0.926 & 0.927 & 0.910 & 0.909 \\
\bottomrule
\end{tabular}
\end{table}

\subsection{Hierarchical Heads versus a Direct Three-Class Model}
\label{sec:threeclass}
The two-head decomposition of Section~\ref{sec:method} is a design choice, so we test
it against the obvious alternative: a single softmax over the three classes, trained on
the same windows with the same encoder. Table~\ref{tab:threeclass} reports the result.
Overall three-class accuracy is $89.0\%$, but the per-class breakdown is what matters.
PD is recovered almost perfectly ($100.0\%$) because it is the majority class, HC
follows at $85.2\%$, and the differential-diagnosis class is by far the hardest at
$80.8\%$. The hierarchical heads reach $93.2\%$ and $90.9\%$ on their respective
binaries, so routing each subject to the head that is meaningful for them is worth
several accuracy points over asking one classifier to separate all three groups at
once. The three-class view is nevertheless the more honest description of the clinical
problem: it shows that the difficulty lives in the other-disorders class, exactly the
group that the pooled binary metrics flatter.

\begin{table}[t]
\caption{Direct three-class HC/PD/DD head (window level, mean over folds). Overall
accuracy is high, but the other-disorders class is recovered far less reliably than
PD, which the pooled binary metrics conceal.}
\label{tab:threeclass}
\centering
\begin{tabular}{lcccc}
\toprule
& \textbf{Overall} & \textbf{HC} & \textbf{PD} & \textbf{DD} \\
\midrule
Three-class accuracy & 0.890 & 0.852 & 1.000 & \textbf{0.808} \\
\bottomrule
\end{tabular}
\end{table}

\subsection{Comparison with a Time-Series Foundation Model}
We benchmarked TimesFM~\cite{das2024}, a decoder-only foundation model pre-trained on
a large time-series corpus, in two regimes (Table~\ref{tab:fm}).
Parameter-efficient fine-tuning with a LoRA adapter~\cite{hu2022}
($r{=}6,\alpha{=}12$) on its hidden state gives the best window-level scores in the
study ($0.977$/$0.969$ AUROC). Used \emph{zero-shot}, however---a frozen feature
extractor with a logistic-regression probe---it reaches only $87.6\%$/$84.4\%$
accuracy and $0.940$/$0.910$ AUROC, below even the compact interleaved encoder
trained from scratch. General time-series pre-training therefore does not by itself
capture PD-specific structure; clinically specific adaptation is required to
discriminate look-alike movement disorders.

\begin{table}[t]
\caption{Foundation-model comparison (window-level, $5$-fold mean). A fine-tuned
foundation model is strongest, but its zero-shot form lags a compact encoder trained
from scratch.}
\label{tab:fm}
\centering
\begin{tabular}{lcccc}
\toprule
& \multicolumn{2}{c}{\textbf{HC vs.\ PD}} & \multicolumn{2}{c}{\textbf{PD vs.\ DD}}\\
\cmidrule(lr){2-3}\cmidrule(lr){4-5}
\textbf{Approach} & Acc & AUROC & Acc & AUROC \\
\midrule
Interleaved self-attn (ours) & 0.932 & 0.963 & 0.909 & 0.960 \\
TimesFM + LoRA         & 0.926 & \textbf{0.977} & 0.913 & \textbf{0.969} \\
TimesFM zero-shot      & 0.876 & 0.940 & 0.844 & 0.910 \\
\bottomrule
\end{tabular}
\end{table}

\subsection{On-Device Latency}
\label{sec:edge}
With minimal preprocessing overhead the deployed model sustains $124.5$~ms per
window---only $6.2\%$ of the $2$~s real-time budget---leaving ample headroom for
continuous on-device monitoring. This efficiency follows directly from the fusion
choice. The interleaved encoder issues a single self-attention operation per block over
the concatenated $2T$ sequence---three multi-head attention calls in total, against the
twelve of explicit cross-attention---and uses $159{,}556$ parameters versus
cross-attention's $418{,}180$ ($2.6\times$ fewer), at no cost in accuracy
(Table~\ref{tab:fusion}). Measured over the same eleven streamed recordings, explicit
cross-attention costs $260.1$~ms per window, roughly $2.1\times$ slower, while late
concatenation costs $119.6$~ms and a single-wrist encoder $75.2$~ms, quantifying the
price of bilateral sensing. On-device power consumption was not measured and is left to
future work.

\subsection{Limitations}
Our study uses a single dataset (PADS); validation on an independent, multi-site cohort
is necessary before deployment, and the patient-disjoint partition must be preserved
throughout to avoid leakage. Self-supervised pre-training is performed on the same data
used for fine-tuning, which likely understates the benefit of contrastive learning; a
fairer test would pre-train on a large external corpus of unlabelled IMU signals. PADS
lacks uniform severity and time-since-diagnosis labels, so we make no claim about
disease staging---only categorical detection and differentiation. Patient-level metrics
saturate at $1.000$ on this cohort, so window-level results carry all of the
discriminative signal reported here and patient-level numbers should not be read as
evidence of clinical readiness. Finally, predictions
have not been clinically verified against expert ground truth beyond the dataset labels.

\section{Conclusion}
We presented a compact interleaved self-attention encoder for Parkinson's disease
detection and, more importantly, PD-versus-DD differential diagnosis from bilateral
wrist-worn IMU signals, under a strict patient-disjoint protocol with threshold-free
metrics. It reaches $0.963$/$0.960$ AUROC, but the low PD-vs-DD sensitivity ($0.812$)
marks the differential as the genuinely hard, clinically decisive boundary for future
work. Our controlled comparison shows that the bilateral \emph{fusion mechanism} is not
what drives performance: interleaved self-attention, explicit cross-attention, late
concatenation, and even single-wrist encoders agree to within $0.001$ AUROC, so we
adopt the interleaved design for costing $2.6\times$ fewer parameters and running
$2.1\times$ faster on device rather than for any accuracy advantage. What does separate
models is the encoder \emph{family}, and only under cross-task generalization---attention
holds $0.94$--$0.96$ AUROC while the CNN collapses and the BiLSTM becomes far more
variable. Signal conditioning (the $0.1$--$20$~Hz band-pass) is the dominant
preprocessing factor, self-supervised pre-training outperforms supervised training by
${\sim}10$ points at a $5\%$ label budget, a foundation model needs fine-tuning to be
competitive, a direct three-class head exposes the other-disorders class as the hardest
to recover ($80.8\%$), and the model runs in real time on a Raspberry~Pi~4. Taken
together, the levers that matter here are signal conditioning, encoder family, and
label budget---not architectural wiring. Future work should target the PD-vs-DD
boundary with severity-annotated cohorts and validate on independent, multi-site data.


\end{document}